\documentclass[letterpaper, 10 pt, conference]{ieeeconf}  % Comment this line out if you need a4paper

\IEEEoverridecommandlockouts                              % This command is only needed if 
\usepackage{url,hyperref,lineno,microtype}
\newcommand{\vect}[1]{\boldsymbol{#1}}
\usepackage{systeme}
\usepackage{gensymb}
\usepackage{amsfonts}
\usepackage{amsmath}
\mathchardef\mhyphen="2D % Define a "math hyphen"
\usepackage{float}
\usepackage{siunitx}
\usepackage{textcomp}
\usepackage[ruled]{algorithm2e}
\usepackage{pbox}
\usepackage{graphicx}
\usepackage{multirow, multicol}

\usepackage[usenames,dvipsnames]{xcolor}

\usepackage{makecell}
\usepackage{pbox}
\usepackage{diagbox}

\usepackage[normalem]{ulem}
\newcommand{\replace}[2]{%
\ifthenelse{ \equal{#1}{} }{}{\textcolor{red}{\sout{#1}}}%
\ifthenelse{ \equal{#2}{} }{}{ \textcolor{olive}{#2}}%
} 

\title{\LARGE \bf
GelTip: A Finger-shaped Optical Tactile Sensor for Robotic Manipulation
}

\author{Daniel Fernandes Gomes$^{1}$, Zhonglin Lin$^{2}$ Shan Luo$^{1}$% 
\thanks{$^{1}$smARTLab, Department of Computer Science, University of Liverpool, Liverpool, United Kingdom. Emails: {\tt\small \{danfergo, shan.luo\}@liverpool.ac.uk} 
$^{2}$School of Mechanical Engineering and Automation, Fuzhou University, Fuzhou, China.}%
}

\begin{document}

\maketitle
\thispagestyle{empty}
\pagestyle{empty}

%%%%%%%%%%%%%%%%%%%%%%%%%%%%%%%%%%%%%%%%%%%%%%%%%%%%%%%%%%%%%%%%%%%%%%%%%%%%%%%%
\begin{abstract}

Sensing contacts throughout the fingers is an essential capability for a robot to perform manipulation tasks in cluttered environments. However, existing tactile sensors either only have a flat sensing surface or a compliant tip with a limited sensing area. In this paper, we propose a novel optical tactile sensor, the~\textit{GelTip}, that is shaped as a finger and can sense contacts on any location of its surface. The sensor captures high-resolution and color-invariant tactile images that can be exploited to extract detailed information about the end-effector's interactions against manipulated objects. Our extensive experiments show that the GelTip sensor can effectively localise the contacts on different locations of its finger-shaped body, with a small localisation error of approximately 5 mm, on average, and under 1 mm in the best cases. The obtained results show the potential of the GelTip sensor in facilitating dynamic manipulation tasks with its all-round tactile sensing capability. The sensor models and further information about the \textit{GelTip} sensor can be found at \textit{http://danfergo.github.io/geltip}.
\end{abstract}

%%%%%%%%%%%%%%%%%%%%%%%%%%%%%%%%%%%%%%%%%%%%%%%%%%%%%%%%%%%%%%%%%%%%%%%%%%%%%%%%
\section{INTRODUCTION}

For both humans and robots, touch provides crucial information about the surfaces under contact. Such information can be used not only to perceive the properties of the contacted objects, such as texture and softness, but to guide the hand motions in manipulation tasks as well. While contacts can happen throughout the entire robot body, in this work we focus on contacts around the gripper fingers, as these are more actively exposed during manipulation. We can group such contacts into ones happening outside or inside of the grasp closure: The former are essential to infer the properties of an object to be grasped, or to detect collisions when approaching the object; The latter are of utter importance to sense the object already being grasped. Hence, to detect both contacts, the development of a tactile sensor that is capable of detecting contacts throughout the entire fingertip surface is of high importance to address robotic manipulation.

\begin{figure}[t]
  \centering
  \includegraphics[width=0.96\linewidth]{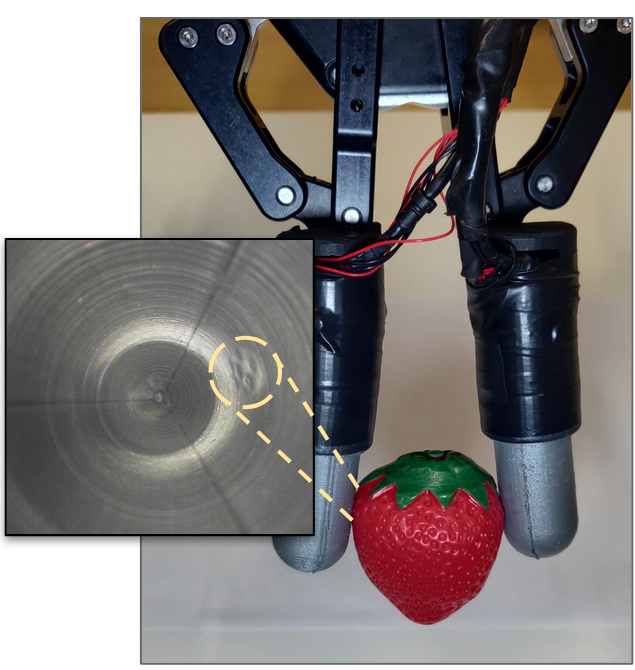} \\
  \caption{A plastic strawberry being grasped by a parallel gripper equipped with two GelTip sensors --- the optical tactile sensor introduced in this paper. Left: In gray-scale, the captured tactile image and corresponding imprint.}
  \label{fig:geltip_sensor}
\end{figure}

Thanks to the use of cameras, optical tactile sensors provide high-resolution images of the deformation caused by contacts with objects in hand. They usually consist of three main parts: A soft elastomer that deforms to the shape of the object upon contact; A webcam underneath that views the deformed elastomer; LEDs that illuminate the space between the elastomer and the webcam. There are two main families of optical tactile sensors, TacTip sensors~\cite{TacTipFamily} and GelSight sensors~\cite{GelSight2017}. TacTip tracks markers printed on a soft domed membrane, while GelSight exploits colored illumination and photometric stereo analysis to reconstruct the membrane deformations. Because of the different working mechanisms, TacTip only measures the surface on a few points, whereas GelSight makes use of the full resolution provided by the camera. However, to the best of our knowledge, only ones that have limited contact measurement areas on one side of the sensor have been proposed.

\begin{figure}
\centering
\includegraphics[width=.99\linewidth]{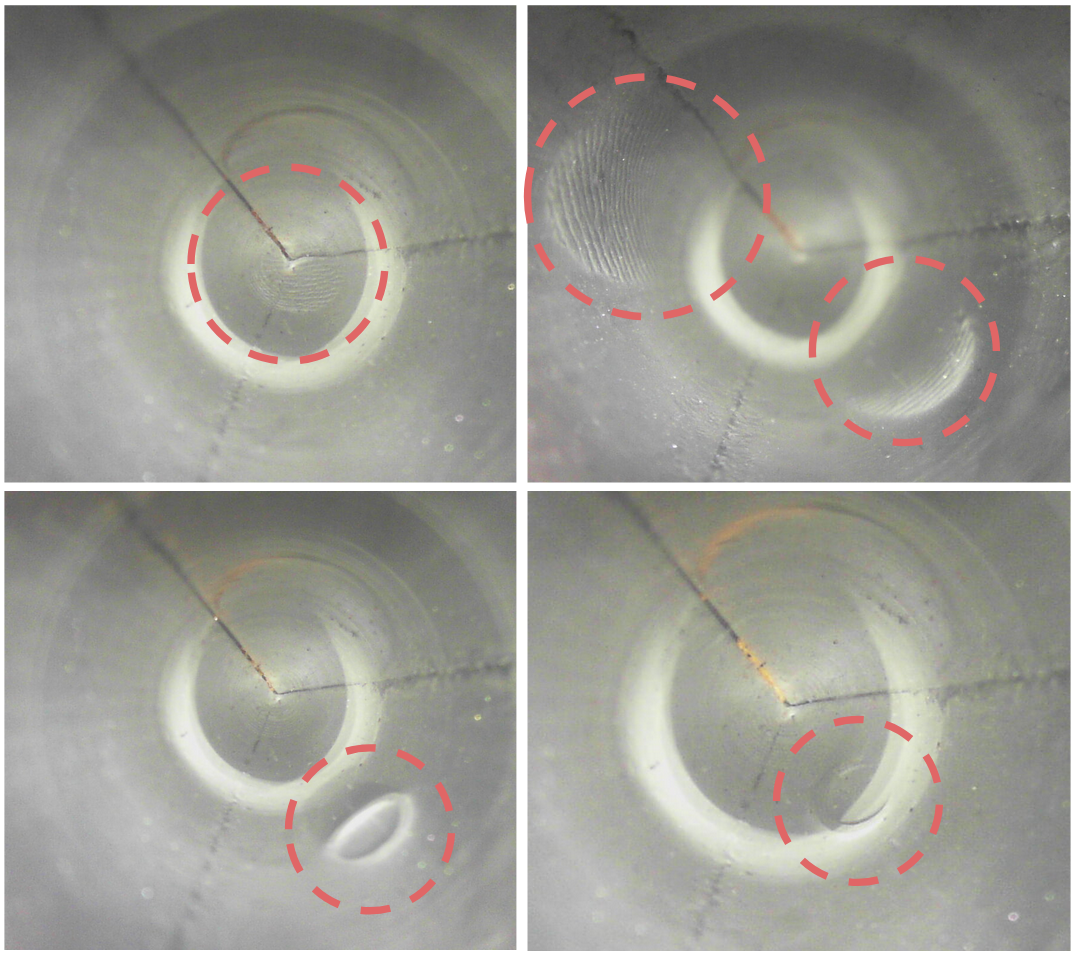} \\
\caption{Tactile images captured using our proposed GelTip sensor. From left to right, top to bottom: a fingerprint pressed against the tip of the sensor, two fingerprints on the sides, a circle being pressed against the side and finally the same circle being pressed against the corner of the tip.}
\label{fig:samples}
\end{figure}

To leverage the full resolution of cameras as GelSight, and to enable the sensor to detect contacts from all the directions, we propose the novel optical tactile sensor \textit{GelTip} that is shaped as a finger and can measure contacts on any location of its surface, using a camera installed at its base, as illustrated in Fig.~\ref{fig:geltip_sensor} and Fig.~\ref{fig:samples}. When  an object is pressed against the tactile membrane, the elastomer distorts and indents the object shape, thus the contacts can be perceived by tracking the changes in the high-resolution outputs of the camera. In contrast with other camera-based tactile sensors, our proposed GelTip sensor is able to detect contacts from a variety of directions, including the frontal and side surfaces, like our human finger. Our extensive experiments show that our proposed GelTip sensor can effectively localise the contacts at different locations of the finger body, with a small localisation error of 5 mm, on average, and under 1 mm in the best cases.

\section{RELATED WORK}
\label{sec:relatedwork}
Compared with remote sensors like cameras, tactile sensors are designed to assess properties of objects such as geometry and texture via physical interactions. A large range of working principles have been actively proposed in the literature~\cite{dahiya2013directions,luo2017robotic}. In this section, we compare the related works in electronic tactile skins and optical tactile sensors that have been widely used for robotic manipulation.

\subsection{Electronic tactile skins}
The electronic tactile skins can be grouped into five categories based on their sensing principles \cite{DexterousTactileSensorsSurvey}: resistive, capacitive, piezoelectric, optical and Organic Field-Effect Transistors (OFETs). These families of tactile sensors measure the pressure distribution of contact by the transduction of a specific electrical characteristic in response to the applied pressure on the surface of the tactile sensor. 
Compared to camera based optical tactile sensors, electronic tactile skins have lower thickness and a smaller size, and can adapt to body parts that have various curvatures and geometry shapes. However, each sensing element of most the tactile skins (for example, a capacitive transducer) has the size of a few square millimetres or even centimetres, which resulted in a limited spatial resolution of the tactile skins. For instance, a commercial Weiss WTS tactile sensor of a similar size to one adult human fingertip has only 14x6 taxels (tactile sensing elements)~\cite{luo2015novel,luo2019iclap}. In addition, they suffer from complicated electronics and cross-talk problems between neighbour sensing elements~\cite{xie2013fiber}.

\subsection{Optical tactile sensors}
Camera based optical tactile sensors make use of cameras to capture touch information. These cameras are placed at the core of an enclosed shell, pointing to a transparent-opaque window made of either soft or two-layered rigid-soft materials. Such characteristics ensure that the captured image is not affected by the external illumination variances. To extract tactile information from the captured images, two main working principles have been proposed: \textit{marker tracking} and \textit{raw image analysis}.

One example of marker tracking-based tactile sensors are the TacTip family: TacTip, TacTip-GR2, TacTip-M2, and TacCylinder~\cite{TacTipFamily,TacTip2009}. Each TacTip sensor introduces novel manufacturing advancements or surface geometries, however, the same working principle is shared: white pins are imprinted onto a transparent through, and coated black, membrane that can be then tracked using computer vision methods. Semi-opaque markers painted at different depths of the elastomer are proposed in~\cite{ColorMixingTactileSensor}, while fluorescent green particles scattered throughout the elastomer are proposed in \cite{greenDots}.

On the other side of the spectrum, the GelSight sensors, initially proposed in~\cite{RetrographicSensing}, exploit the entire resolution of the tactile images captured by the camera and high accuracy geometry reconstructions can be produced~\cite{GelSightSmallParts,lee2019touching}.
The sensor was equipped to a robotic gripper that inserts a USB cable into the correspondent port~\cite{GelSightSmallParts}. To employ methods that were explored in marker-based sensors, markers were added to the membrane as well~\cite{GelSight2017}. There are also attempts to adapt the morphology of the sensor~\cite{GelSlim}. However, the sensor only measures a small flat area that is oriented towards the grasp closure.

In these previous works, multiple designs and two distinct working principles have been proposed, however, none of the introduced sensors have the capability of sensing the entire surface of a robotic finger, i.e., the sides and the tip. As as result, current experiments in object manipulation leveraging these sensors are highly constrained, with contacts only being detected when the manipulated object is inside the grasp closure~\cite{GelSightSmallParts, IncipientSlip}. To address this gap, we propose the GelTip finger-shaped sensor that is able to detect the contact throughout the sensor surface. 

% ======================================================================================
% ========                             SENSOR MODEL                         ============
% ======================================================================================
\section{The GelTip sensor}
\subsection{Overview}
\label{sec:sensormodel}

\begin{figure}
\centering
\includegraphics[width=.7\linewidth]{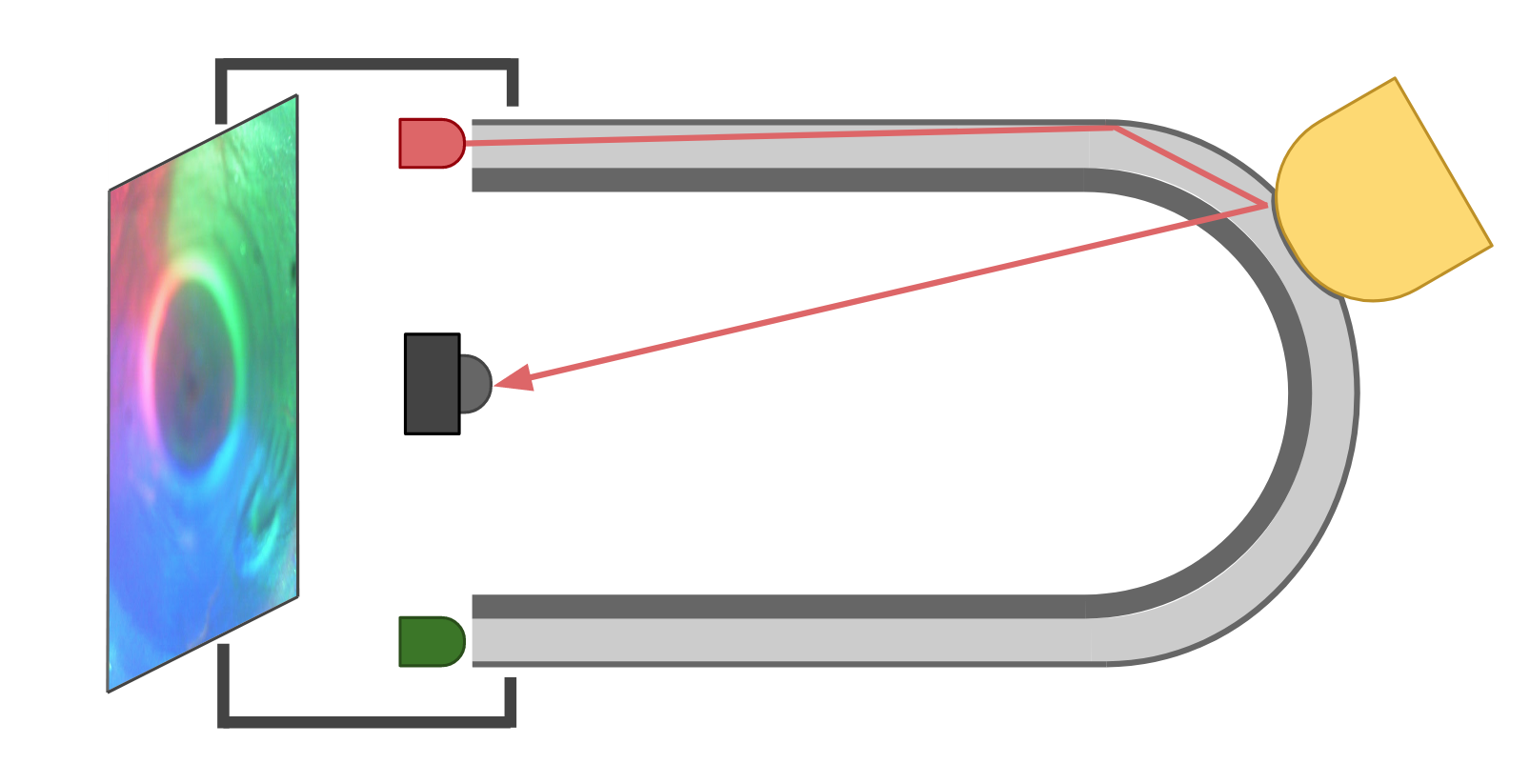} \\
\caption{The working principle of the proposed \textit{GelTip} sensor. The three-layer tactile membrane (rigid body, elastomer and paint coating) is shown in gray. The light rays emitted by the LEDs travel through the elastomer. As an object, shown in yellow, presses the soft elastomer against the rigid body, an imprint is generated and an image is captured by the camera placed in the core of the tactile sensor. An opaque shell that encloses all the optical components ensures the constant internal lighting of the elastomer surface. }

\label{fig:sensor_innerworkings}
\end{figure}

As illustrated in Fig.~\ref{fig:sensor_innerworkings}, the introduced GelTip body consists of three layers, from the inside to the outer surface: a rigid transparent body, a soft transparent membrane and a layer of opaque elastic paint. A camera is placed at the base of the sensor, looking from the inside of the tube. When an object is pressed against the tactile membrane, the elastomer distorts and indents the object shape. The camera captures the obtained imprint into a digital image. Since one property of tactile sensing is being immune to external light variations, the camera is enclosed within an opaque shell, with the tactile membrane being the only interface with the external environment. Thanks to the finger shape of the sensor, the LED light sources can be placed adjacent to the base of the sensor and the strategically controlled light rays are guided through the tube and elastomer. 

\hfill \break
\subsection{Surface projection into the tactile image}
\label{subsec:sensor_geometry}

The tactile images captured by the sensor can be processed to retrieve information about the contacted surfaces, i.e., the object geometry, the contact location and force distributions etc. Methods for obtaining such information have been introduced in previous works~\cite{GelSight2017,GelSightSmallParts}. However, due to the flat surface of the existing GelSight sensors, the relationship between the camera and the elastomer surfaces has not been explicitly considered in previous works. Here we derive the projective function, $m$, that maps pixels in the image space $(x',y')$ into points $ (x,y,z)$ on the sensor surface. Such a projection model is necessary for, among other applications, detecting the contact locations on the 3D sensor surface.

\begin{figure}
  \centering
  \includegraphics[width=.8\linewidth]{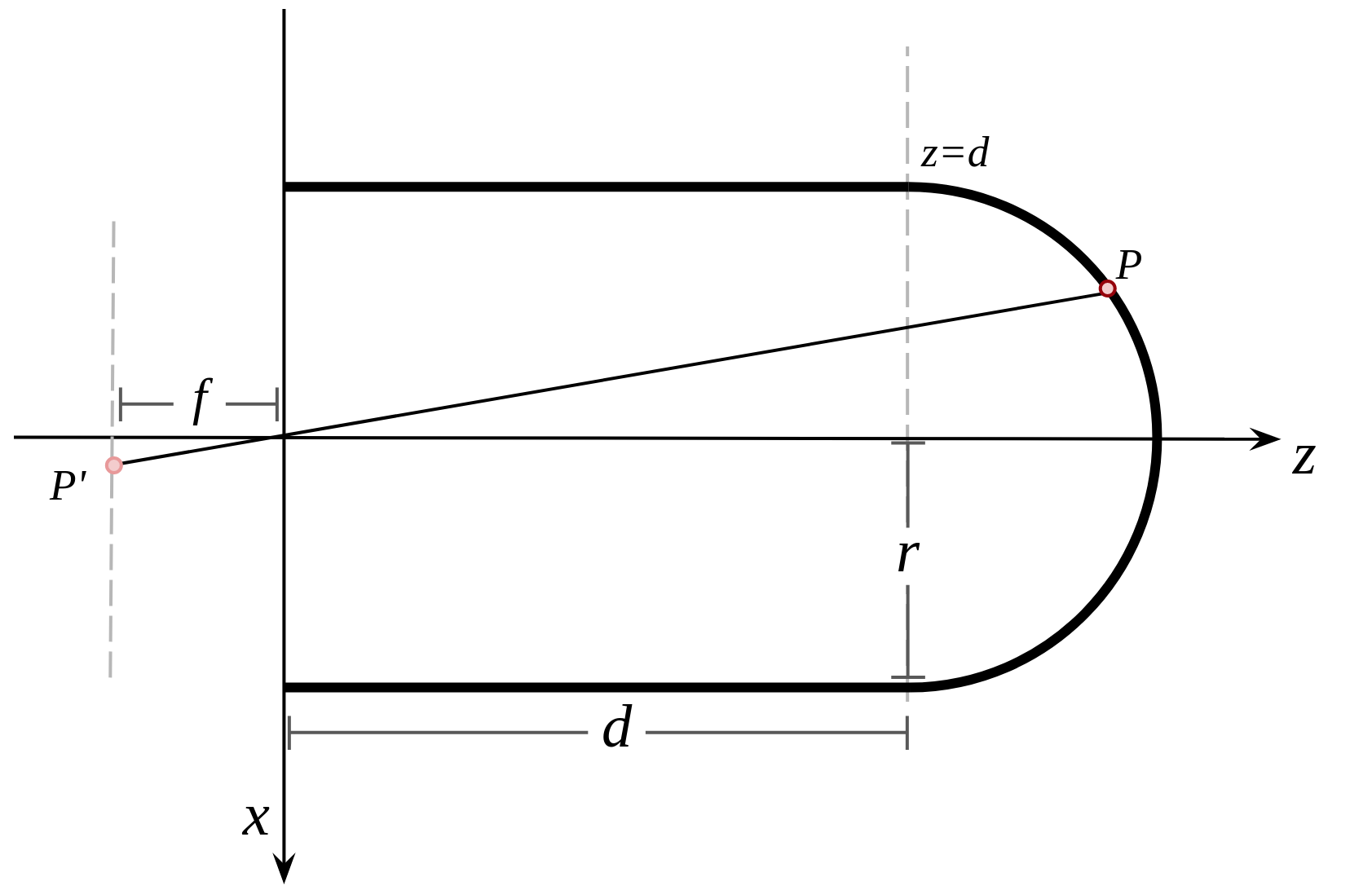}
  \caption{Two-dimensional representation of the \textit{GelTip} sensor geometrical model. The tactile membrane is modeled as a cylindrical surface and a semi-sphere. An optical sensor with its focal point placed in the frame of reference origin and with a focal-length $f$, projects a point $P$ on the sensor surface, into $P'$ in the image plane. }
  \label{fig:sensor_geometry}
\end{figure}

As shown in Fig.~\ref{fig:sensor_geometry}, the sensor surface can be modeled as a joint semi-sphere and an open cylinder, both sharing the same radius $r$. The cylinder surface center axis and the $z$-axis are collinear, therefore, the center point of the semi-sphere can be set to $(0,0,d)$, where $d$ is the distance from the center point of the base of the semi-sphere to the center point of the base of the sensor. The camera is oriented in the direction of the $z$ axis, with its focal point placed at the referential origin $(0,0,0)$. As a result, the image space is centered around the $z$ axis. The location of any point on the sensor surface $(x,y,z)$ can be represented as follows:

\begin{align}
x^2 + y^2 + (z - d)^2 &= r^2,    &&  z > d \\
\label{eq:surface_cylinder}
x^2 + y^2 &= r^2,                && z <= d
\end{align}

By making the usual thin lens assumptions, we model the optical sensor as an ideal pinhole camera. The projective transformation that maps a point in the world space $ P $ into a point in the tactile image $P'$ can be defined using the general camera model~\cite{szeliski2011computer} as:
\begin{align}
\label{eq:camera_model}
P' &= K [\vect{R} | \vect{t}] P \\
K &= \begin{bmatrix} 
	\begin{array}{llll}
        fk & 0  & c_x & 0 \\
        0  & fl & c_y & 0 \\
        0  & 0  & 1   & 0 \\
    \end{array}    
%fk\frac{x}{z} + c_x \\
%fl\frac{y}{z} + c_y \\
\end{bmatrix}
\end{align}
where \mbox{$P'=[x'z , y'z, z]^T$} is an image pixel and \mbox{$P=[x,y,z,1]^T$} is a point in space, both represented in homogeneous coordinates here. $[R|t]$ is the camera's extrinsic matrix that encodes the rotation $ R $ and translation $ t $ of the camera, $K$ is the camera intrinsic matrix ($f$ is the focal length; $k$ and $l$ are the pixel-to-meters ratios; $c_x$ and $c_y$ are the offsets in the image frame). Assuming that the camera produces square pixels, i.e., $k = l$, $fk$ and $fl$ can be replaced by $\alpha$, for mathematical convenience.

The orthogonal projections on $XZ$ and $YZ$ of a given projection ray are obtained by expanding the matrix multiplication given by Eq.~\ref{eq:camera_model} and solving it \mbox{w.r.t. $x$ and $y$}:

\begin{align}
\systeme*{
x'z = \alpha x + c_xz, 
y'z = \alpha y + c_yz,
z = z
} \Leftrightarrow
\systeme*{
\alpha x = x'z - c_xz,
\alpha y = y'z - c_yz
} \Leftrightarrow
\systeme*{
x = (\frac{x' - c_x}{\alpha})z,
y = (\frac{y' - c_y}{\alpha})z
}
\label{eq:proj_ray}
\end{align}

The desired mapping function $m: (x',y') \rightarrow (x,y,z) $ can then be obtained by constraining the $z$ coordinate through the intersection of the generic projection ray with the sensor surface: 

\begin{equation}
\left\{
\begin{array}{ll}
		
z &= \left\{
\begin{array}{lll}
\sqrt{\frac{(r\alpha)^2}{({x' - c_x})^2 + ({y' - c_y})^2}} \\[4pt]
\hspace{52pt}  \mbox{if } ({x' - c_x})^2 + ({y' - c_y})^2  < (\frac{r\alpha}{d})^2 \\
\quad \\ 
\frac{{\alpha^2}2d \pm \sqrt{{(-{\alpha^2}2d)}^2 - 4{[{(x' - c_x)}^2 + {(y' - c_y)}^2 ]}{[{(d^2 - r^2)}{\alpha^2}]}}}{2{[{(x' - c_x)}^2 + {(y' - c_y)}^2 + \alpha^2]}}\\
\hspace{150pt} \mbox{otherwise}
\end{array}
\right. \\
\quad \\
x &= (\frac{x' - c_x}{\alpha})z \\
y &= (\frac{y' - c_y}{\alpha})z
\end{array}
\right.
\label{eq:function_f}
\end{equation}

\begin{figure}
  \centering
  \includegraphics[width=.8\linewidth]{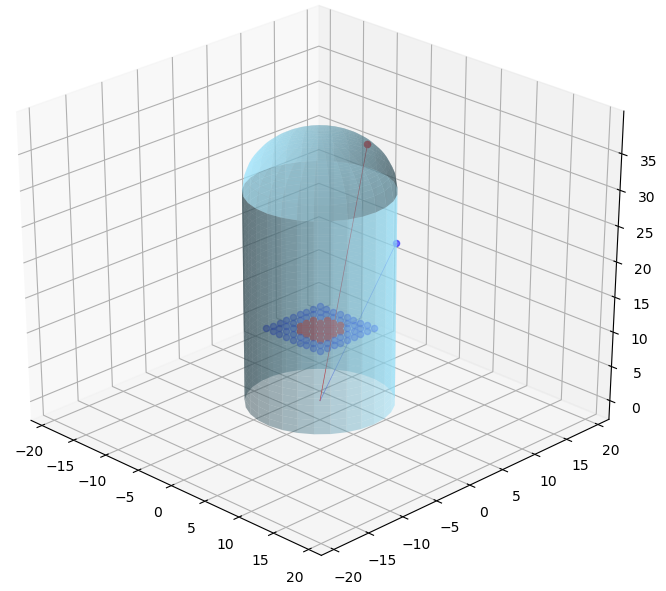} \\
  \caption{Two projection rays that correspond to the spherical (in red) and cylindrical (in navy blue) regions are depicted in the figure. Each ray intersects three relevant points: the frame of reference origin, a point in the sensor surface and the corresponding projected point in the image plane.}
  \label{fig:sim_model}
\end{figure}

The introduced sensor model is validated and visualised in Fig.~\ref{fig:sim_model}.

\subsection{Fabrication process}
\label{subsec:fabrication_process}

The major challenges to address when producing a compact finger-shaped visuo-tactile sensor, arise from its compact size and the non-flat surface, namely the fabrication of the tactile membrane.

An off-the-shelf transparent test tube is used to construct the rigid layer of the sensor body, simplifying (or avoiding) the fabrication of the necessary curved surface. One disadvantage of using the off-the-shelf test tubes, particularly the plastic ones, is that these contain imperfections.

The remaining necessary rigid parts to build the sensor body are: A \textit{shell}, where the camera electronics and LEDs are installed; A \textit{sleeve} that is glued onto the test tube and tightened to the \textit{shell}; a \textit{supporting base} that secures the sensor into the gripper finger; And a three-part \textit{mold} used to cast the elastomer. To fabricate the rigid parts, we take advantage of 3D printing technology. We experiment with printing the parts using both  Fused Filament Fabrication (FFF) and Stereolithography (SLA) printers, i.e., the Anycubic i3 Mega and the Formlabs Form 2. The models and further information about the GelTip sensor are available at \textit{http://danfergo.github.io/geltip}.
 
We use the same materials as suggested in \cite{GelSight2017}, i.e., XP-565 from Silicones, Inc. (High Point, NC, USA) and Slacker® from Smooth-on Company. After extensive experiments we find the best ratios to be 1:22:22, i.e., 1 gram of XP-565 part-A, 22 grams of XP-565 part-B and 22 grams of the Slacker. This amount of mixture is sufficient to fabricate two sensor membranes. The ratio \textit{part-A/part-B}  influences the rigidity of the elastomer, i.e., higher concentration of \textit{part-B} produces a softer silicone. The Slacker, on the other hand, contributes to the silicone tackiness. It is necessary to add sufficient Slacker to make the elastomer capable of capturing high frequency imprints such as a fingerprint. However, Slacker makes the silicone sticky if too much is added.

After the elastomer is cured and de-molded, we proceed with painting. Off-the-shelf spray paints tend to form a rigid coat and cracks will develop in the coat when the elastomer deforms or stretches. To avoid these issues, we apply a custom paint coating using the airbrush method suggested in \cite{GelSight2017}. We mix the coating pigment with a small portion of \textit{part-A} and \textit{part-B} of XP-565, with the same ratio used in the elastomer. We experiment with both the Silver Cast Magic® from the Smooth-on Company and the aluminium powder (\SI{1}{\micro\metre}) from the US Research Nanomaterials, Inc. After mixing them properly, we dissolve the mixture using a silicone solvent until we achieve a watery liquid. The liquid paint is then sprayed onto the elastomer surface using an airbrush. 

Three sets of LEDs are then soldered; either of different colors, red, green and blue, or all white. They are inserted into the three corresponding pockets in the sensor \textit{sleeve}. Since different LEDs emit different light intensities, we solder each cluster to a different wire and resistor before connecting them to the power source. The values of these resistors are manually tuned and vary from \SI{30}{\ohm} to \SI{600}{\ohm}. The power source can be either extracted from the camera USB cable, by splicing it, or adding a secondary USB cable. 

\begin{figure}
  \centering
  \includegraphics[width=.9\linewidth]{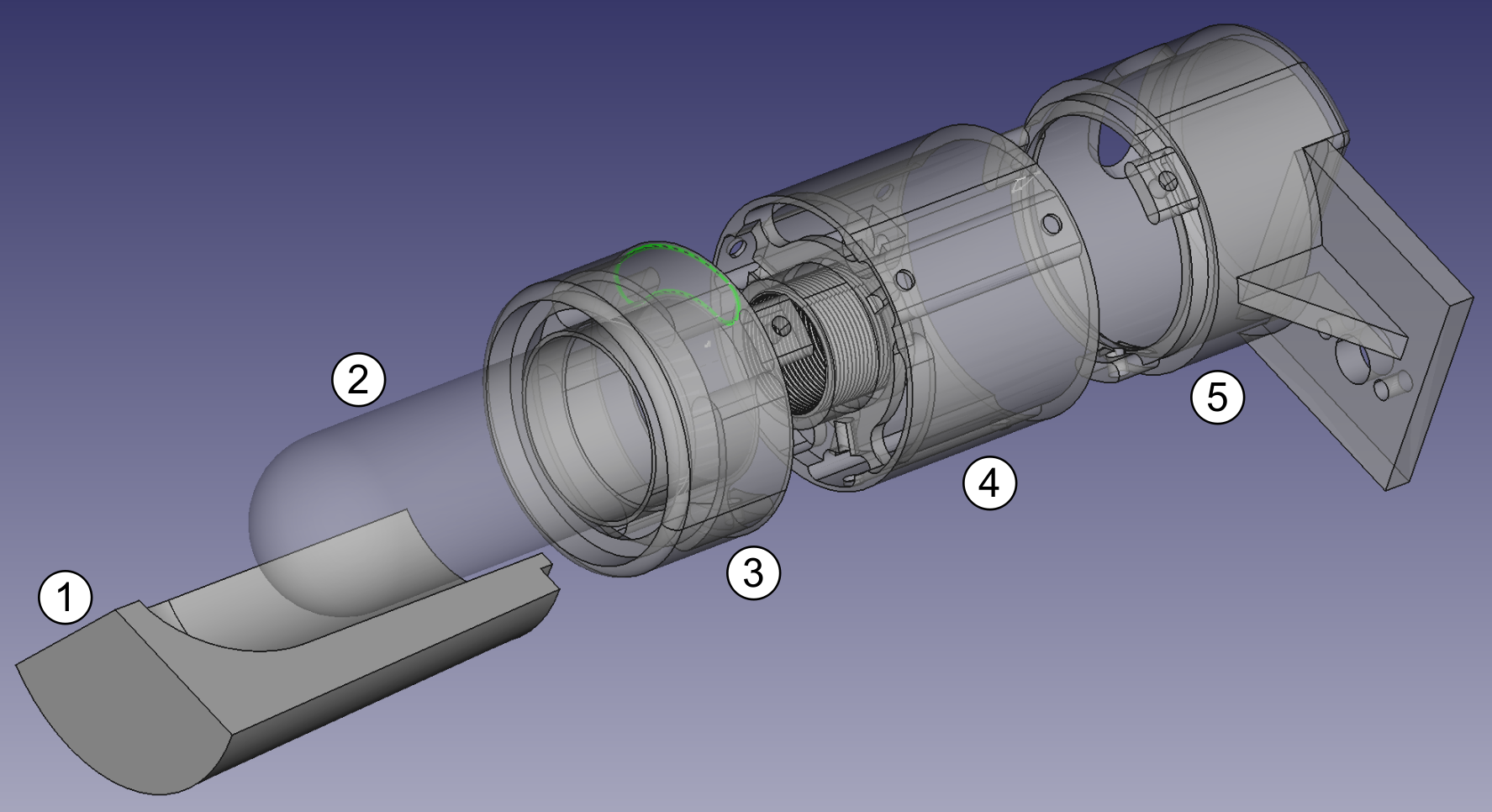} \\
  \caption{Exploded view of the GelTip tactile sensor design. From left to right,  (1) one of three parts of the elastomer mold; (2) a representation of the plastic test tube;  (3) the \textit{sleeve} that secures the tactile membrane; (4) the \textit{shell} to hold the M12 lens, the camera electronics and the LEDs; and (5) the sensor \textit{supporting base}, containing the main electronics.}
  \label{fig:sensor_cad}
\end{figure}

\section{Experiments}
\label{sec:experimentresults}

To validate the design of our sensor we carry a set of experiments that aim to demonstrate that the GelTip sensor can effectively localise the contacts at different locations of its finger-shaped body, and consequently it can help a robot carrying out of manipulation tasks.

\subsection{Sensor construction parameters}

In section \ref{sec:sensormodel},  the GelTip generic model and fabrication process is described, however some parameters, that highly affect the quality of the obtained tactile image are not discussed. In this section we summarise the results of the experiments carried out to decide such parameters.

\paragraph{Sensor radius and lens viewing angle}

One key parameter to consider when designing the GelTip sensor, is its tactile membrane radius. Considering a fixed length in the $z$ direction, the radius influences not only the sensor compactness but its lateral observable area as well. This happens due to the fact the projection rays are oblique, and thus the larger the tube the further away they intersect it, resulting in a smaller observable side area. We experiment with three differently sized tubes, i.e., \SI{21}{\mm}, \SI{15}{\mm} and \SI{13}{\mm}; and find that the \SI{15}{\mm}, being the smallest tube that enables us to fit the camera lens inside, offers an intermediate compromise between observable side area, the sensor compactness and flexibility for adjusting the camera position (depth-wise). With regard to the camera lens and its viewing angle, we experiment with the default Microsoft LifeCam lens (70\degree) and an off-the-shelf M12 wide-lens (170\degree). We find that given the tube size configuration, the default 70\degree ~viewing angle is insufficient to capture the sides of the finger and, thus the 170\degree~wider lens is necessary. 

\paragraph{Painting and illumination}

\begin{figure}
  \setlength{\tabcolsep}{0pt}
  \centering
  \begin{tabular}[b]{c}
    \includegraphics[width=0.47\linewidth]{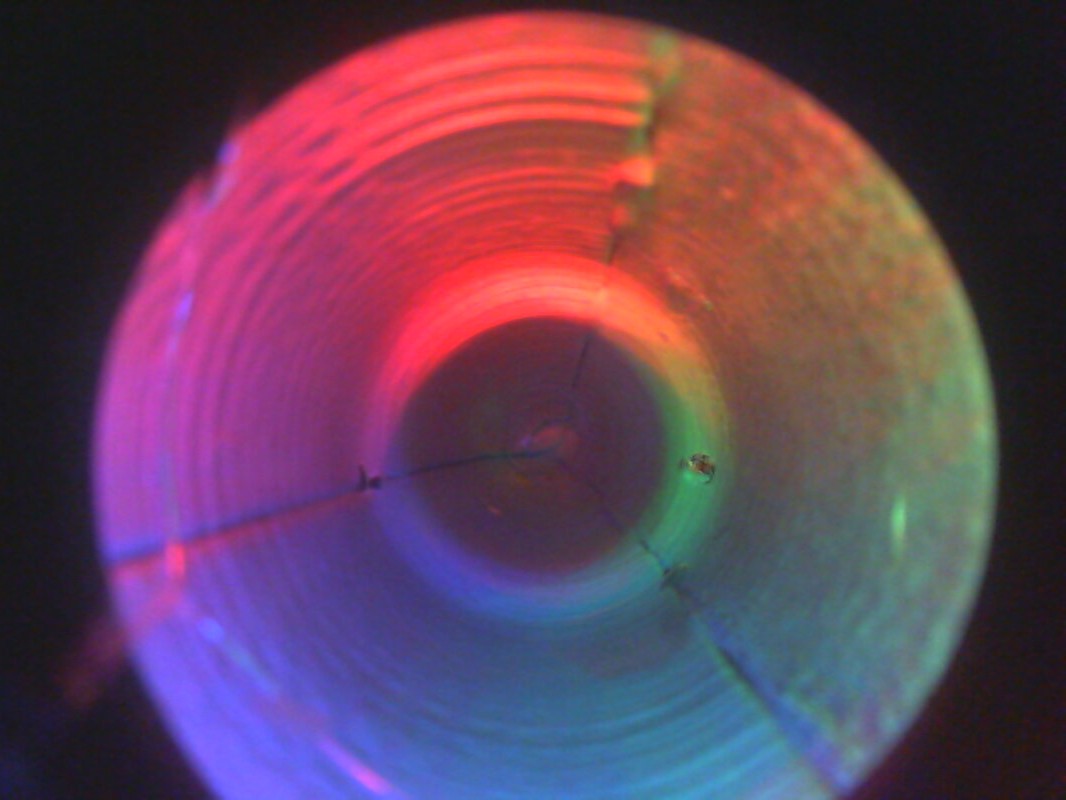} \\
    % \small (a) Aluminium-powder paint
  \end{tabular}
  \begin{tabular}[b]{c}
    \includegraphics[width=0.47\linewidth]{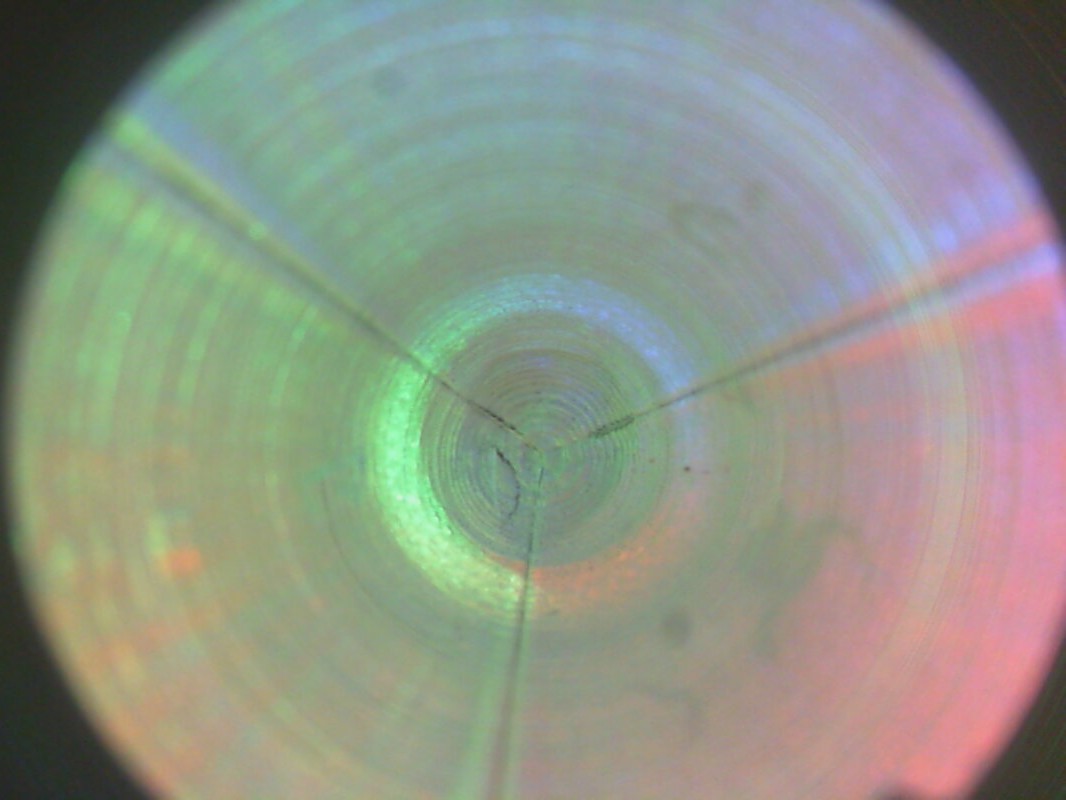} \\
    % \small (b) Off-the-shelf elastic paint 
  \end{tabular}
  \caption{Comparison of two differently painted tactile membranes. On the left, painted using aluminium powder based paint, and, on the right, off-the-shelf metallic elastic paint. It should be noted the darker central region, in the aluminium powder painted surface.}
  \label{fig:paints}
\end{figure}

In \cite{GelSight2017}, aluminium powder paint is suggested to reduce the existence of specular reflections, and consecutively improve the image quality for surface reconstruction. In our experiments, we find that due to the curved surface of the GelTip, this powder results in an highly non uniform color distribution. Further, the tip of the finger is poorly illuminated, resulting in a darker central region and poor tactile signal when contacts are here applied. This effect is depicted in Fig.~\ref{fig:paints}, wherein we compare two views of differently painted tactile membranes: aluminium powder based paint and off-the-shelf metallic elastic paint. 

\begin{figure}
\setlength{\tabcolsep}{0pt}
  \centering
  \begin{tabular}[b]{c}
    \includegraphics[width=0.47\linewidth]{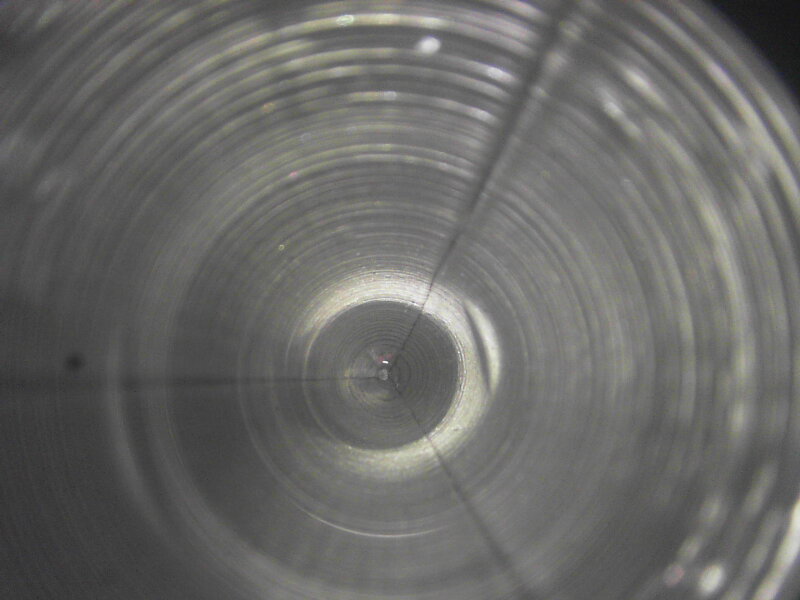} \\
    % \small (a) Aluminium-powder paint
  \end{tabular}
  \begin{tabular}[b]{c}
    \includegraphics[width=0.47\linewidth]{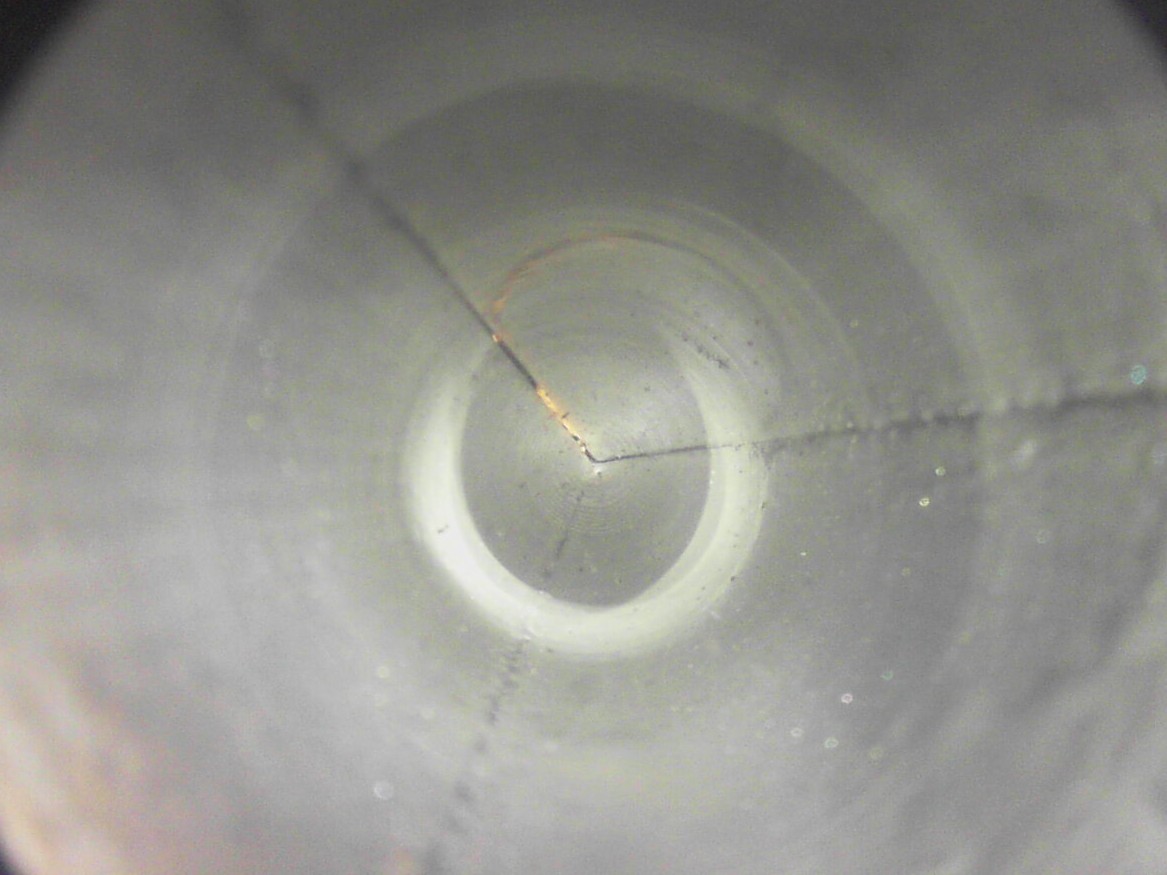} \\
    % \small (b) Off-the-shelf elastic paint 
  \end{tabular}
  \caption{Comparison of surface textures obtained using two different the two 3D printing technologies: Fused Filament Fabrication (FFF), on the left, and Stereolithography (SLA), on the right. }
  \label{fig:roughness}
\end{figure}

\paragraph{Surface roughness}

In \cite{GelSlim}, the usage of textured fabric is discussed as an approach for increasing the responsiveness of the tactile signal. We experiment with printing the necessary molds for shaping the elastomer using Fused Filament Fabrication (FFF) and Stereolithography (SLA) printers. The different obtained surfaces are shown in Fig.~\ref{fig:roughness}. The texture caused by FFF printing is particularly noticeable closer to the sensor base, being possibly detrimental for the recognition of high frequency geometries, such as fingerprints.

% ==============================================================
% ================================ CONTACT DETECTION
% ==============================================================

\subsection{Contact localisation}
\label{subsec:contact_detection}

\begin{figure}
\centering
\includegraphics[width=1\linewidth]{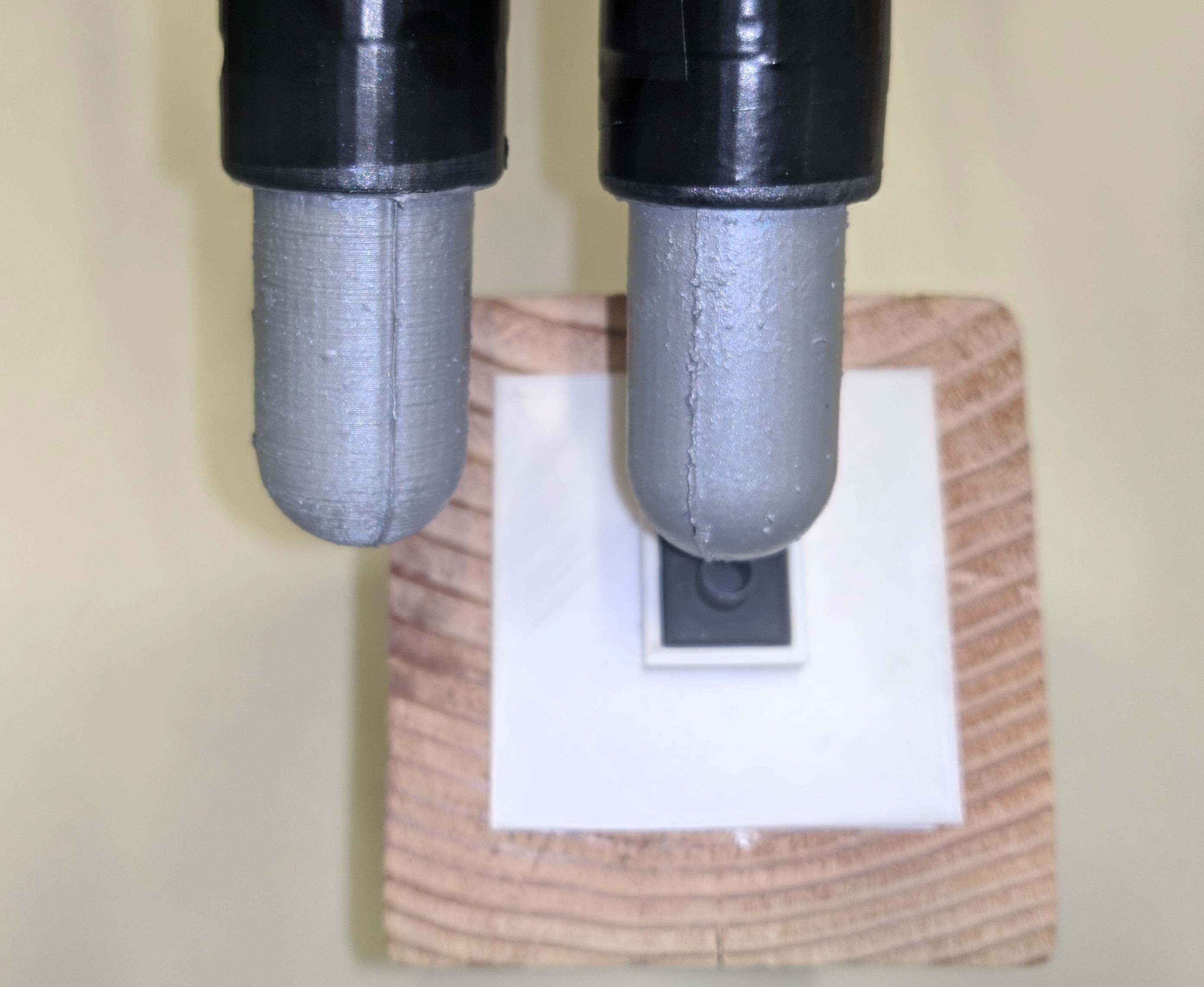} \\
\caption{Two GelTip sensors are installed on a robotic actuator and a 3D printed mount is placed on top of a wooden block holding small 3D printed solid. e.g. a cylinder. The actuator moves in small increments and collects tactile images annotated with the known contact position.}
\label{fig:exp_contact_detection}
\end{figure}

The main argument behind proposing a finger shaped optical tactile sensor is to be able to detect contacts throughout the entire finger surface. To evaluate this capability, a GelTip sensor is  installed  on a robotic actuator, and a 3D printed mount is placed on top of a wooden block, holding one of the following small 3D-printed solids: cone, sphere, irregular prism, cylinder, edge, tube or slab; as shown in Fig.~\ref{fig:exp_contact_detection}. The actuator moves and taps these objects in known positions. A total of eight taps are carried out using four different orientations around the fingertip: $0$, $\pi/6$, $\pi/4$ and $\pi/3$; and four translations on its side: \SI{0}{\mm}, \SI{5}{\mm}, \SI{10}{\mm}, and \SI{15}{\mm}. The robot tapping motion starts with the gripper pointing downwards i.e., orientation $0$, and ends with the sensor in an horizontal position, contacting the object \SI{15}{\mm} from the discontinuity region, i.e. translation \SI{15}{\mm}.

\begin{figure}
\centering
\includegraphics[width=1\linewidth]{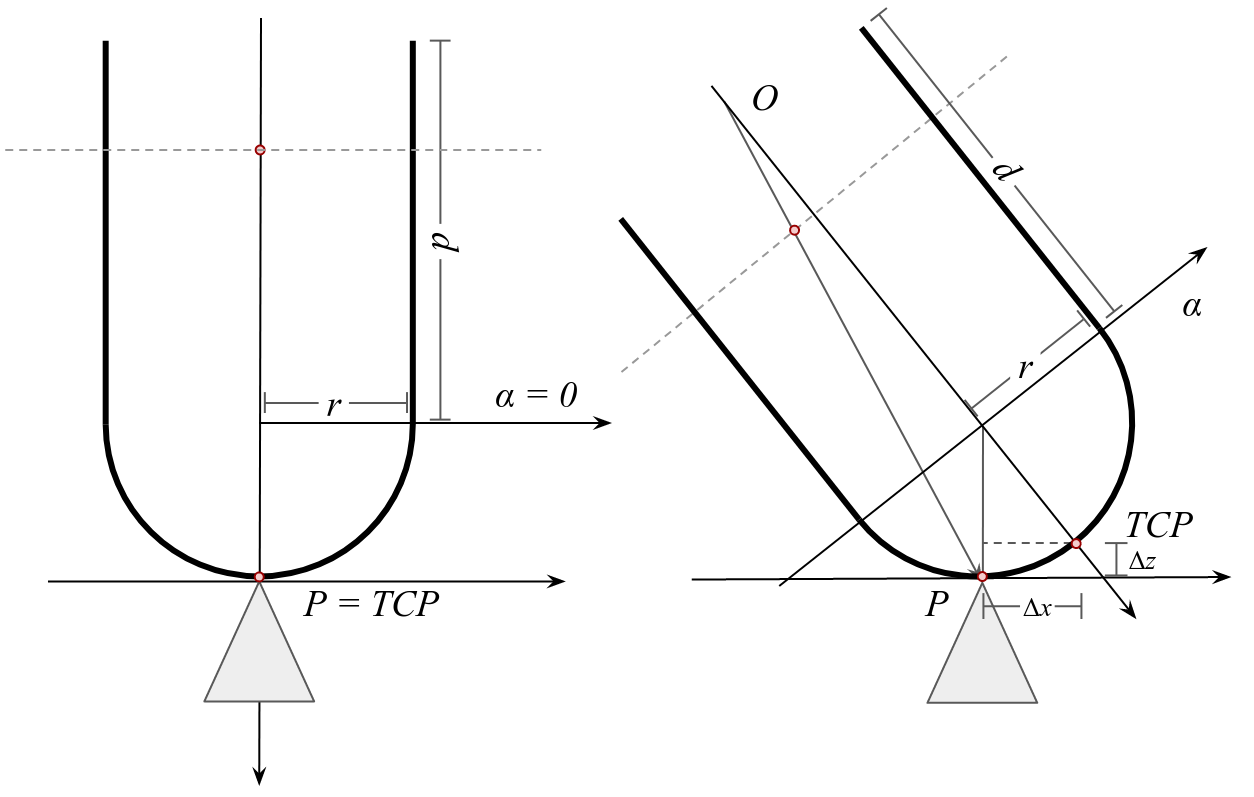} \\
\caption{The sensor is set pointing downwards and centered using the captured tactile image. We then move the in small first longitudinal and then rotational increments. To compensate for the finger thickness and obtain a contact on the finger skin, we adjust its final position by $\Delta x $ $\Delta z$.}
\label{fig:exp_motion_model}
\end{figure}

Initially, the actuator is set pointing downwards and visually aligned with the tip of the first object, the cone. The actuator Tool Center Point (TCP) is set as the sensor tip, and consequently the first contact position is the initial TCP position. For the remaining positions, we add the corresponding orientation and translation to the initial one. To compensate the finger thickness, and to obtain contacts on the finger skin, we adjust its position by a small increment, as shown in Fig.~\ref{fig:exp_motion_model}. 

To automatically detect the contacts, a simple image subtraction based algorithm is implemented. Before each contact, a reference image is also captured. When a contact occurs, the absolute difference between the reference and the in-contact frames is computed. The attained difference frame is then filtered using a mean convolution ($15\times15$  kernel), and pixels with a brightness intensity lower than 60\% are set to zero. Then, using the OpenCV \emph{findContours} function, in-contact regions are determined. These regions are further filtered by the size of their area i.e., only clusters with an area between 0.012\% and 0.04\% of the total picture area are kept. The center point of such regions is then found using the OpenCV \emph{fitEllipse} function. A final prediction is set as the weighted average of clusters centers. 

To map the detected contacts into positions on the surface of the finger, the projection model described in Section~\ref{sec:sensormodel} is used. The model has five parameters: $r$, $d$, $c_x$, $c_y$ and $\alpha$. The first two, $r$ and $d$ are extracted from the dimensions of the sensor design, however, the latter three are the intrinsic parameters of the camera, which need to be calibrated. To this end, the $c_x, c_y$ parameters are manually annotated with the first contact. To obtain $\alpha$, the \SI{15}{\mm} translation contact is used. The contact is annotated in image space and the obtained pair $P' \rightarrow P$ is fit into Eq.~\ref{eq:proj_ray} to derive $\alpha$. 

Finally, the Euclidean distance between the predicted and true contact positions is measured. We execute one run (8 contacts) per each of the 7 objects, and the aggregate results are reported in Table~\ref{table:contact_errors_per_position} and Table~\ref{table:contact_errors_per_object}.  
Overall, the variance between the observed and true localisation errors is substantial, in some contacts the obtained errors are around \SI{1}{\mm}, while in others are over \SI{1}{\cm}. On the other hand, the localisation error, for each object or position, is correlated with its variance. The largest localisation errors happen on objects with large or rounded tops i.e., sphere, edge and slab; contrariwise, the lowest errors are observed for objects with sharp tops, i.e., cone, tube and cylinder. In terms of the localisation errors at different positions, contacts happening near the sensor tip, i.e., the rotations, show lower errors than contacts happening on the sensor side, i.e., translations. In particular, contacts happening at $\pi/4$ and $\pi/6$ have the lowest errors. 

From the data and our observations during the experiment, we conclude that the obtained errors arise from three sources: 1) weak imprints; 2) the flexing of the sensor and 3) imperfections in the sensor modeling and calibration. For instance, localisation errors $> 1cm$ are commonly due to the imperceptible contact imprints, the algorithm incorrectly predicting the contact around the sensor tip.

\begin{table}
\setlength{\tabcolsep}{3.9pt}
\def\arraystretch{2.2}
\centering
\caption{Contact errors per position, expressed in millimeters}
\label{table:contact_errors_per_position}
\begin{tabular}{ccccccccc}
\Xhline{3\arrayrulewidth}
\multicolumn{4}{c}{\textbf{ROTATIONS}} & \multicolumn{4}{c}{\textbf{TRANSLATIONS}} \\[-1.5ex]
$0$ & $\pi/6$ & $\pi/4$ & $\pi/3$ & 0 & 5 & 10 & 15 \\ 
\Xhline{1.5\arrayrulewidth} 
\makecell{$4.71$\\$\pm0.75$} & 
\makecell{$2.01$\\$\pm0.90$} & 
\makecell{$1.04$\\$\pm0.46$} & 
\makecell{$6.96$\\$\pm4.82$} & 
\makecell{$7.87$\\$\pm5.08$} & 
\makecell{$8.03$\\$\pm1.92$} & 
\makecell{$7.55$\\$\pm5.00$} & 
\makecell{$4.86$\\$\pm8.41$} \\ 
\Xhline{3\arrayrulewidth} 
\end{tabular}
\end{table}

\begin{table}
\setlength{\tabcolsep}{5.2pt}
\def\arraystretch{2.2}
\centering
\caption{Contact errors per object, expressed in millimeters}
\label{table:contact_errors_per_object}
\begin{tabular}{cccccccccc}
\Xhline{3\arrayrulewidth}
\pbox{0.093\linewidth}{\includegraphics[width=0.95\linewidth]{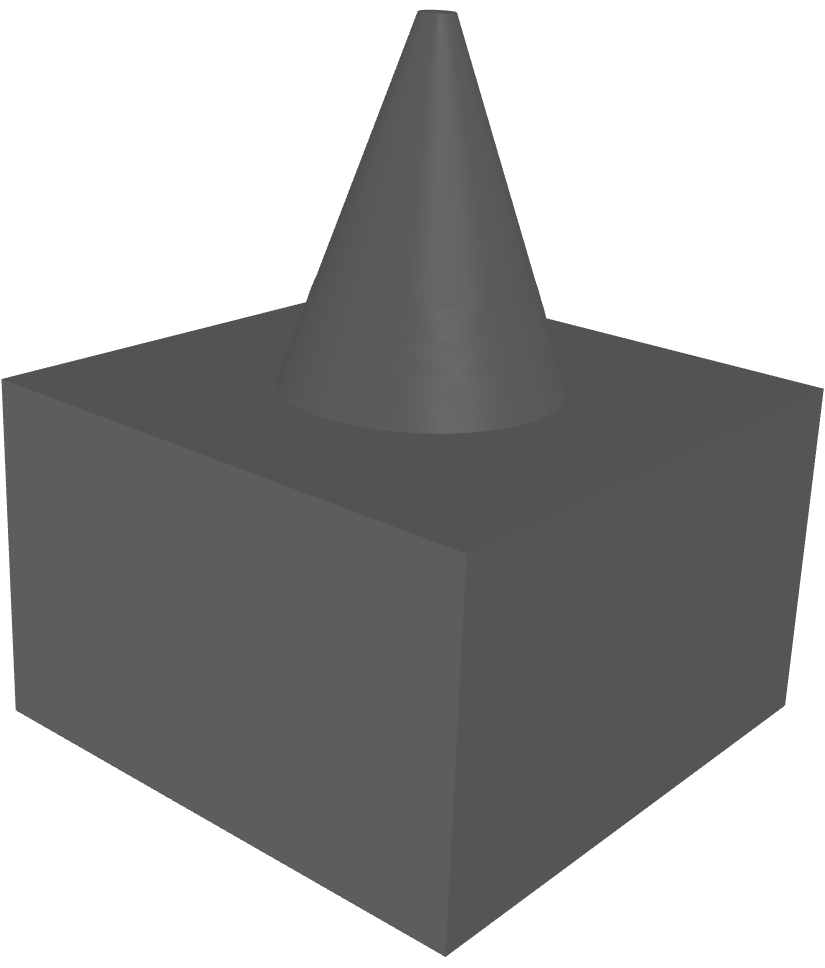} \\ \centering\textbf{Cone} }&%
\pbox{0.096\linewidth}{\includegraphics[width=\linewidth]{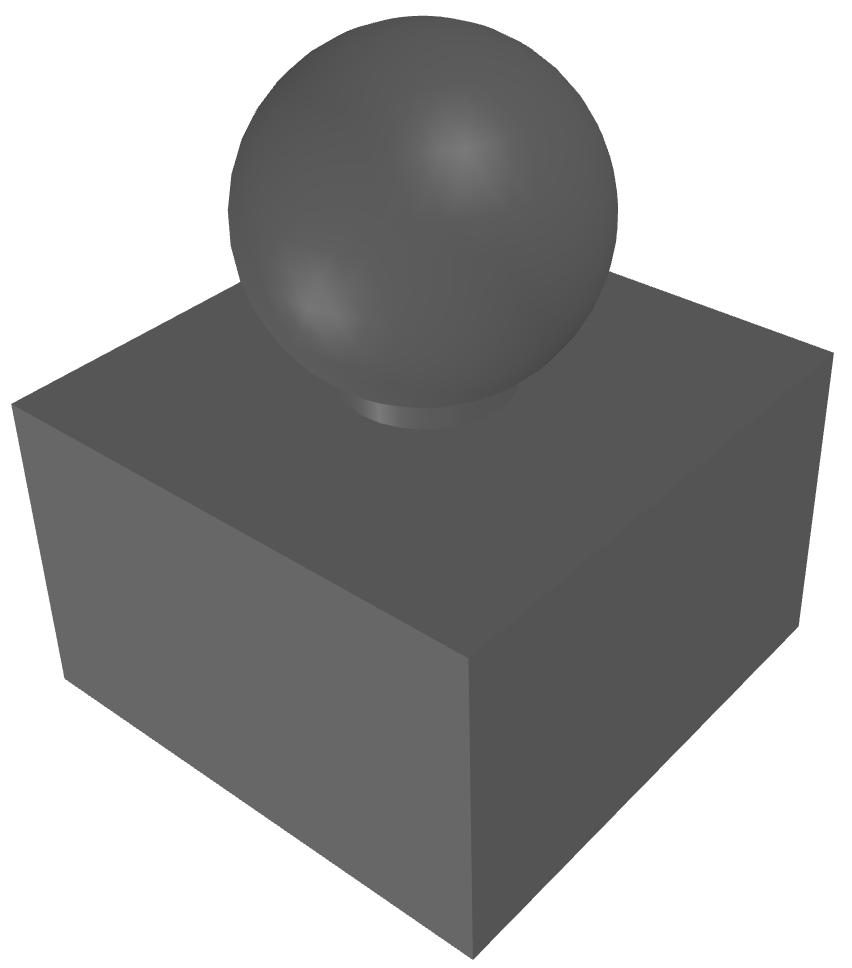} \\ \centering\textbf{Sphere} }&%
\pbox{0.12\linewidth}{\includegraphics[width=0.8\linewidth]{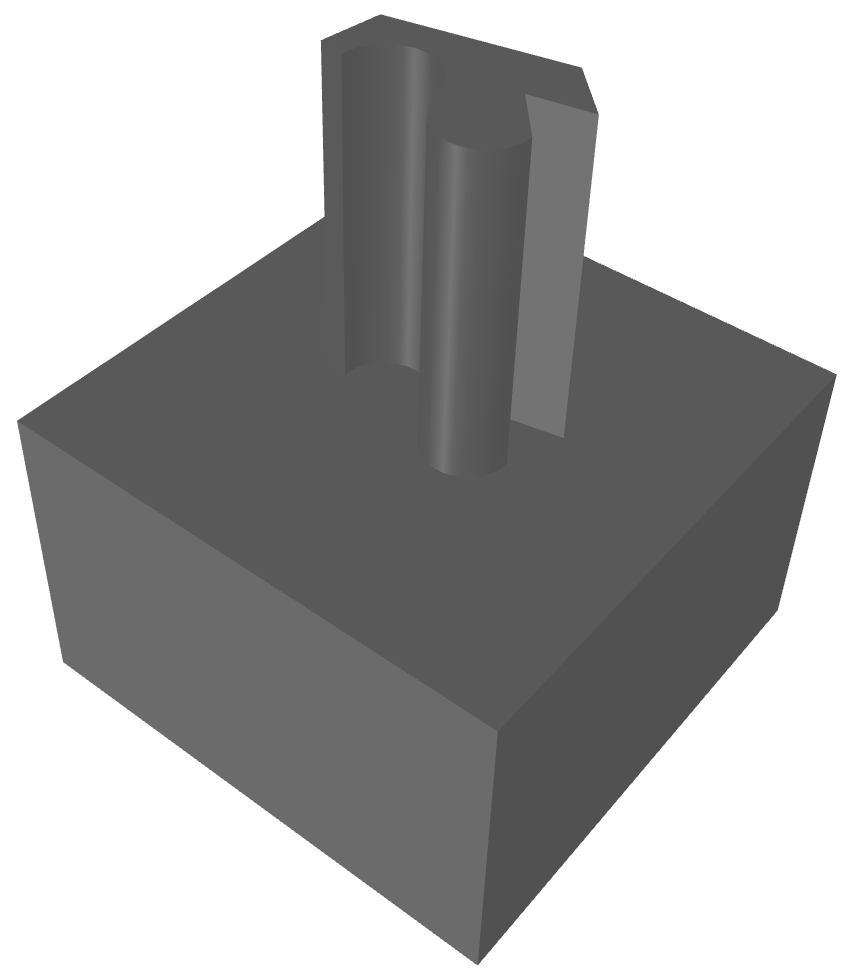} \\ \centering\textbf{Irregular} }&%
\pbox{0.094\linewidth}{\includegraphics[width=\linewidth]{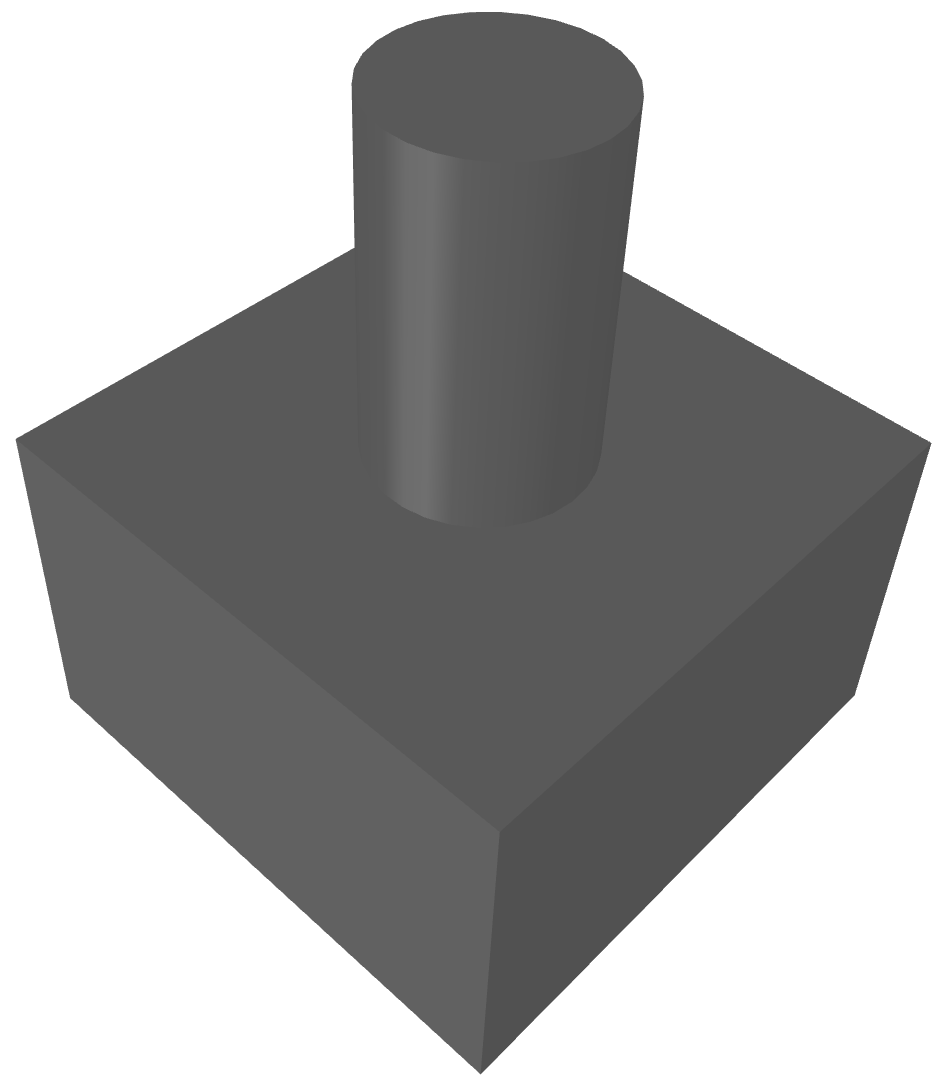} \\ \centering\textbf{Cylinder} }&%
\pbox{0.093\linewidth}{\includegraphics[width=\linewidth]{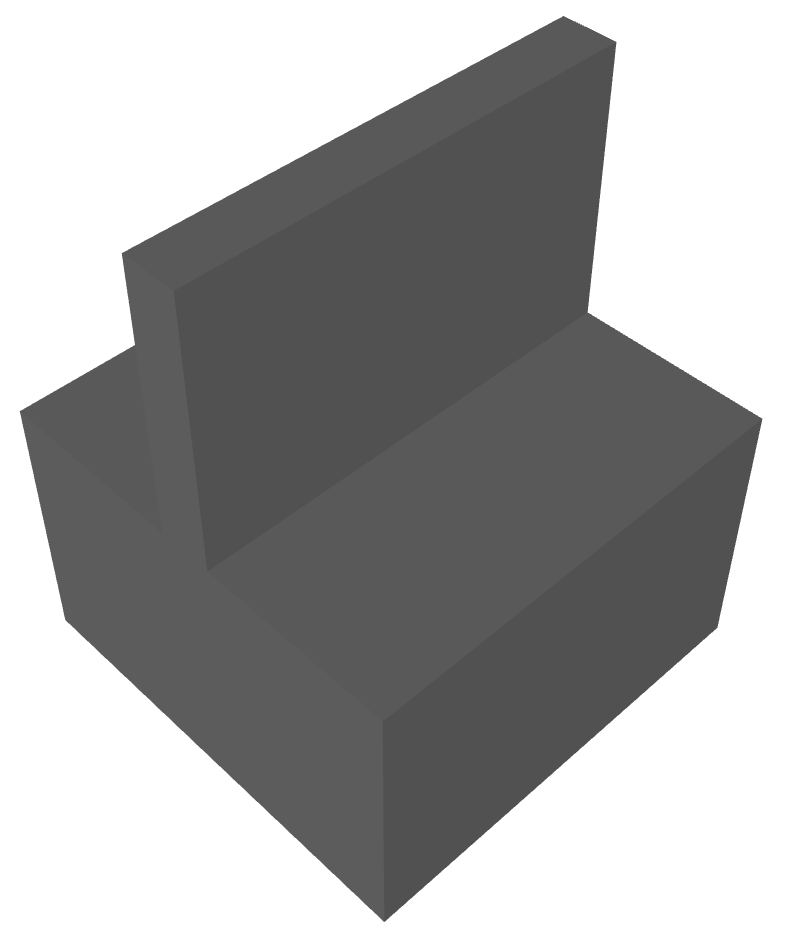} \\ \centering\textbf{Edge} }&%
\pbox{0.093\linewidth}{\includegraphics[width=\linewidth]{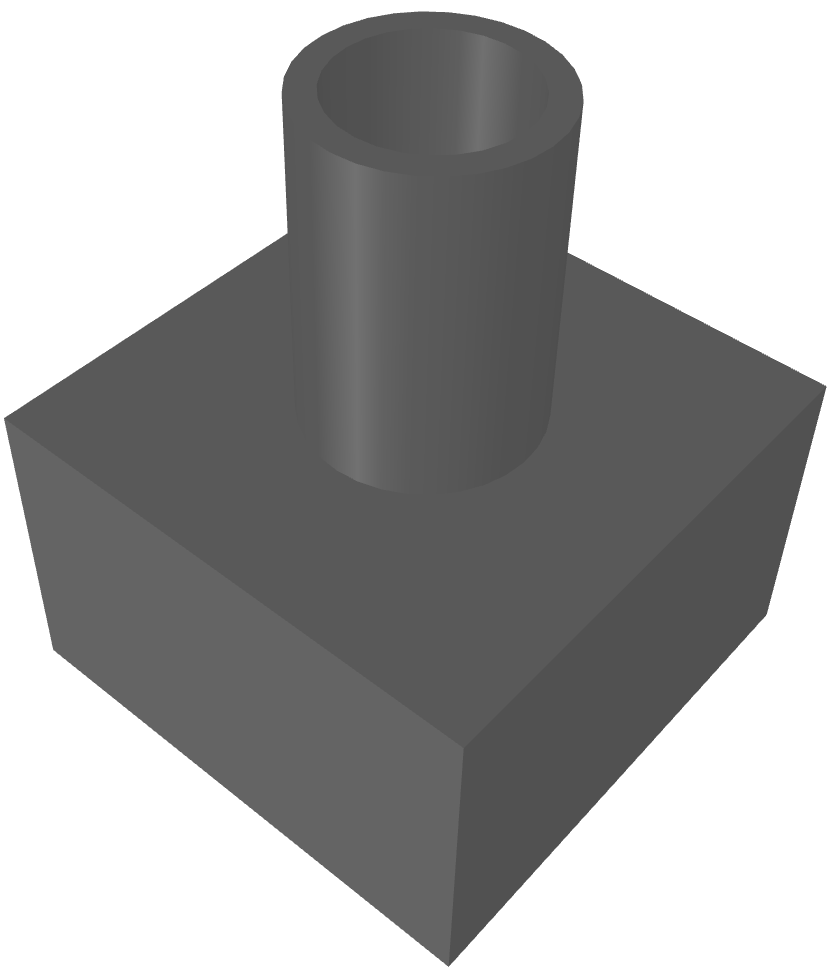} \\ \centering\textbf{Tube} }&%
\pbox{0.093\linewidth}{\includegraphics[width=0.9\linewidth]{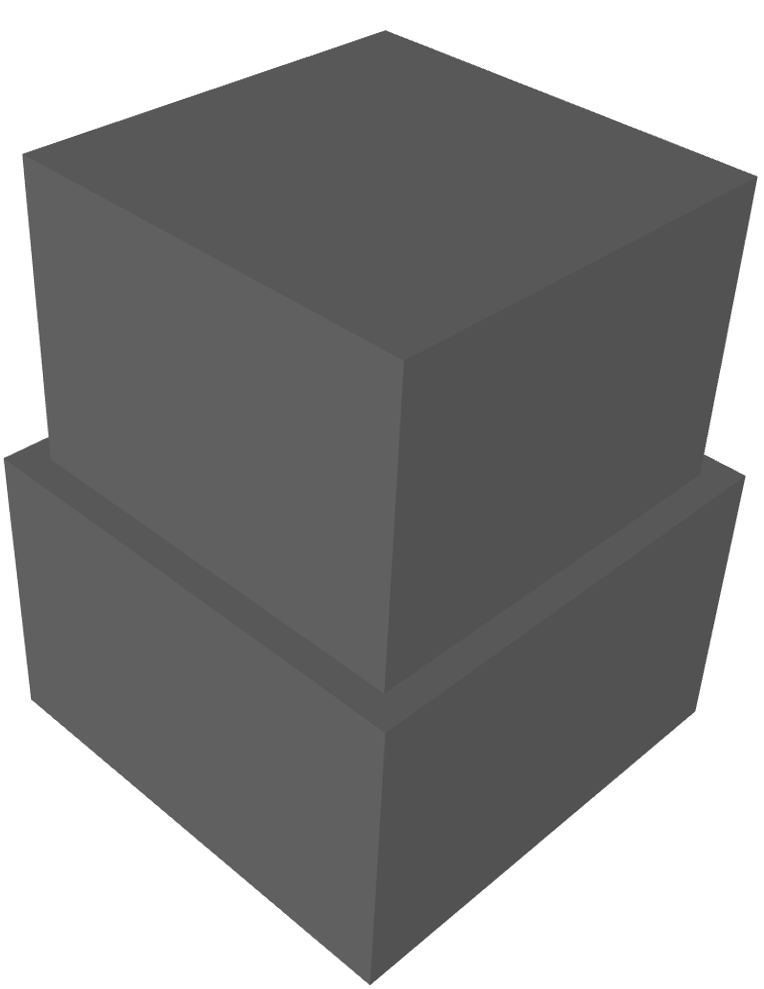} \\ \centering\textbf{Slab} } \\ 
\Xhline{1.5\arrayrulewidth} 
\makecell{$3.63$\\$\pm3.26$} & 
\makecell{$6.79$\\$\pm5.38$} & 
\makecell{$5.61$\\$\pm4.08$} & 
\makecell{$4.57$\\$\pm4.30$} & 
\makecell{$7.47$\\$\pm6.29$} & 
\makecell{$3.33$\\$\pm1.90$} & 
\makecell{$6.27$\\$\pm8.17$} \\
\Xhline{3\arrayrulewidth} 
\end{tabular}
\end{table}

\section{CONCLUSIONS}
\label{sec:conclusion}
In this paper, the GelTip optical tactile sensor is proposed. The introduced sensor offers multiple advantages when compared against other camera-based tactile sensors, namely the fact that the GelTip is able to capture high-resolution readings throughout its entire finger-shaped surface. Our experiments show that, for instance, the GelTip sensor can effectively localise these contacts with a  small error  of approximately \SI{5}{\mm}, on average, and under \SI{1}{\mm} in the best cases. While the obtained error is significant for tactile sensing, it should be noted that this error is justified by the simplistic nature of the contact detection algorithm used. From Fig.~\ref{fig:samples}, it can be seen that the sensor can effectively capture much detailed textures, such as human fingerprints. As such, better quantitative results should be obtained with better localisation methods. Furthermore, same as the GelSight sensor, the GelTip sensor is also expected to achieve surface reconstruction \cite{GelSight2017} and incipient split detection \cite{IncipientSlip}.

In the future research, improvements to the sensor design and further works demonstrating the advantages of all-round tactile sensing, should be carried out. We will also use the GelTip sensor in the context of manipulation tasks, such as grasping in cluttered environments or manipulation of deformable objects.

\section*{ACKNOWLEDGMENT}
This work was supported by the EPSRC project ``Robotics and Artificial Intelligence for Nuclear" (EP/R026084/1).

\addtolength{\textheight}{-1cm}   % This command serves to balance the column lengths

%%%%%%%%%%%%%%%%%%%%%%%%%%%%%%%%%%%%%%%%%%%%%%%%%%%%%%%%%%%%%%%%%%%%%%%%%%%%%%%%

%%%%%%%%%%%%%%%%%%%%%%%%%%%%%%%%%%%%%%%%%%%%%%%%%%%%%%%%%%%%%%%%%%%%%%%%%%%%%%%%

%%%%%%%%%%%%%%%%%%%%%%%%%%%%%%%%%%%%%%%%%%%%%%%%%%%%%%%%%%%%%%%%%%%%%%%%%%%%%%%%
%\section*{APPENDIX}

% \section*{ACKNOWLEDGMENT}
% The preferred spelling of the word ÒacknowledgmentÓ in America is without an ÒeÓ after the ÒgÓ. Avoid the stilted expression, ÒOne of us (R. B. G.) thanks . . .Ó  Instead, try ÒR. B. G. thanksÓ. Put sponsor acknowledgments in the unnumbered footnote on the first page.

%%%%%%%%%%%%%%%%%%%%%%%%%%%%%%%%%%%%%%%%%%%%%%%%%%%%%%%%%%%%%%%%%%%%%%%%%%%%%%%%

\bibliographystyle{ieeetr}
\bibliography{root.bib}

\begin{thebibliography}{10}

\bibitem{TacTipFamily}
B.~Ward-Cherrier, N.~Pestell, L.~Cramphorn, B.~Winstone, M.~E. Giannaccini,
  J.~Rossiter, and N.~F. Lepora, ``{The TacTip Family: Soft Optical Tactile
  Sensors with 3D-Printed Biomimetic Morphologies},'' {\em Soft Robotics},
  vol.~5, no.~2, pp.~216--227, 2018.

\bibitem{GelSight2017}
W.~Yuan, S.~Dong, and E.~H. Adelson, ``{GelSight: High-Resolution Robot Tactile
  Sensors for Estimating Geometry and Force.},'' {\em Sensors (Basel,
  Switzerland)}, vol.~17, 11 2017.

\bibitem{dahiya2013directions}
R.~S. Dahiya, P.~Mittendorfer, M.~Valle, G.~Cheng, and V.~J. Lumelsky,
  ``Directions toward effective utilization of tactile skin: A review,'' {\em
  IEEE Sensors Journal}, vol.~13, no.~11, pp.~4121--4138, 2013.

\bibitem{luo2017robotic}
S.~Luo, J.~Bimbo, R.~Dahiya, and H.~Liu, ``Robotic tactile perception of object
  properties: A review,'' {\em Mechatronics}, vol.~48, pp.~54--67, 2017.

\bibitem{DexterousTactileSensorsSurvey}
H.~Yousef, M.~Boukallel, and K.~Althoefer, ``{Tactile sensing for dexterous
  in-hand manipulation in robotics - A review},'' {\em Sensors and Actuators,
  A: Physical}, vol.~167, no.~2, pp.~171--187, 2011.

\bibitem{luo2015novel}
S.~Luo, W.~Mou, K.~Althoefer, and H.~Liu, ``Novel tactile-sift descriptor for
  object shape recognition,'' {\em IEEE Sensors Journal}, vol.~15, no.~9,
  pp.~5001--5009, 2015.

\bibitem{luo2019iclap}
S.~Luo, W.~Mou, K.~Althoefer, and H.~Liu, ``iclap: Shape recognition by
  combining proprioception and touch sensing,'' {\em Autonomous Robots},
  vol.~43, no.~4, pp.~993--1004, 2019.

\bibitem{xie2013fiber}
H.~Xie, H.~Liu, S.~Luo, L.~D. Seneviratne, and K.~Althoefer, ``Fiber optics
  tactile array probe for tissue palpation during minimally invasive surgery,''
  in {\em IEEE/RSJ International Conference on Intelligent Robots and Systems
  (IROS)}, pp.~2539--2544, 2013.

\bibitem{TacTip2009}
C.~Chorley, C.~Melhuish, T.~Pipe, and J.~Rossiter, ``{Development of a tactile
  sensor based on biologically inspired edge encoding},'' {\em 2009
  International Conference on Advanced Robotics, ICAR 2009}, pp.~1--6, 2009.

\bibitem{ColorMixingTactileSensor}
X.~Lin and M.~Wiertlewski, ``{Sensing the Frictional State of a Robotic Skin
  via Subtractive Color Mixing},'' {\em IEEE Robotics and Automation Letters},
  vol.~4, no.~3, pp.~2386--2392, 2019.

\bibitem{greenDots}
C.~Sferrazza and R.~D’Andrea, ``Design, motivation and evaluation of a
  full-resolution optical tactile sensor,'' {\em Sensors}, vol.~19, no.~4,
  p.~928, 2019.

\bibitem{RetrographicSensing}
M.~K. Johnson and E.~H. Adelson, ``Retrographic sensing for the measurement of
  surface texture and shape,'' in {\em 2009 IEEE Conference on Computer Vision
  and Pattern Recognition}, pp.~1070--1077, IEEE, 2009.

\bibitem{GelSightSmallParts}
R.~Li, R.~Platt, W.~Yuan, A.~ten Pas, N.~Roscup, M.~A. Srinivasan, and
  E.~Adelson, ``Localization and manipulation of small parts using gelsight
  tactile sensing,'' in {\em 2014 IEEE/RSJ International Conference on
  Intelligent Robots and Systems}, pp.~3988--3993, IEEE, 2014.

\bibitem{lee2019touching}
J.-T. Lee, D.~Bollegala, and S.~Luo, ``{“Touching to See” and “Seeing to
  Feel”: Robotic Cross-modal Sensory Data Generation for Visual-Tactile
  Perception},'' in {\em ICRA}, 2019.

\bibitem{GelSlim}
E.~Donlon, S.~Dong, M.~Liu, J.~Li, E.~Adelson, and A.~Rodriguez, ``Gelslim: A
  high-resolution, compact, robust, and calibrated tactile-sensing finger,'' in
  {\em 2018 IEEE/RSJ International Conference on Intelligent Robots and Systems
  (IROS)}, pp.~1927--1934, IEEE, 2018.

\bibitem{IncipientSlip}
S.~Dong, D.~Ma, E.~Donlon, and A.~Rodriguez, ``Maintaining grasps within
  slipping bounds by monitoring incipient slip,'' in {\em 2019 International
  Conference on Robotics and Automation (ICRA)}, pp.~3818--3824, IEEE, 2019.

\bibitem{szeliski2011computer}
R.~Szeliski, ``Computer vision algorithms and applications,'' ch.~2, Springer,
  2011.

\end{thebibliography}

\end{document}